%% file: kronberger.tex
\documentclass{elsarticle}

\journal{Applications in Engineering Sciences}

\usepackage{amsmath}
\usepackage{booktabs} 
\usepackage{siunitx}
\usepackage{multirow}
\usepackage{colortbl} 
\usepackage{gensymb} 
\usepackage{tikz} 
\usetikzlibrary{arrows.meta} 

\tikzset{%
  >={Latex[width=2mm,length=2mm]},
         process/.style = {rectangle, rounded corners, draw=black,
                           minimum width=2cm, minimum height=1cm,
                           text centered, font=\sffamily},
}

\begin{document}
\begin{frontmatter}
\title{Extending a Physics-Based Constitutive Model using Genetic Programming}
  \author[jrcaddr]{Gabriel Kronberger}
   \ead{gabriel.kronberger@fh-hagenberg.at}
  
   \author[tumucaddr]{Evgeniya Kabliman}
  \author[lkraddr]{Johannes Kronsteiner}
  \author[jrcaddr]{Michael Kommenda}
  
   \address[jrcaddr]{%
     Josef Ressel Center for Symbolic Regression\\
     University of Applied Sciences Upper Austria\\
     Softwarepark 11, 4232 Hagenberg, Austria
   }
  
   \address[tumucaddr]{%
     Technical University of Munich\\
     TUM School of Engineering and Design\\
     Department of Materials Engineering \\
     Chair of Materials Engineering of Additive Manufacturing\\
     Boltzmannstr. 15, 85748 Garching, Germany
   }
   
   \address[lkraddr]{%
     LKR Light Metals Technologies\\
     Austrian Institute of Technology\\
     Giefinggasse 2, 1210 Vienna, Austria
   }

\begin{abstract}
  In material science, models are derived to predict emergent material properties (e.g. elasticity, strength, conductivity) and their relations to processing conditions. A major drawback is the calibration of model parameters that depend on processing conditions. Currently, these parameters must be optimized to fit measured data since their relations to processing conditions (e.g. deformation temperature, strain rate) are not fully understood. We present a new approach that identifies the functional dependency of calibration parameters from processing conditions based on genetic programming. We propose two (\emph{explicit} and \emph{implicit}) methods to identify these dependencies and generate short interpretable expressions. The approach is used to extend a physics-based constitutive model for deformation processes. This constitutive model operates with internal material variables such as a dislocation density and contains a number of parameters, among them three calibration parameters. The derived expressions extend the constitutive model and replace the calibration parameters. Thus, interpolation between various processing parameters is enabled.
  Our results show that the implicit method is computationally more expensive than the explicit approach but also produces significantly better results. 
\end{abstract}


\end{frontmatter}

\input{kronberger-body}


\end{document}

%% file: kronberger-body.tex
\section{Introduction and Motivation}

Mathematical models are at the core of science and engineering and allow us to predict physical phenomena without
direct observations. Only through modelling and simulation we are able to build extremely complex and safe physical
objects (such as air planes, space vehicles, or power plants). In empirical modelling one can distinguish white-box and
black-box models with a whole spectrum of grey-box models between these two extremes \cite{Sjoberg1995,VonStosch2014}.
White-box models can be derived from physical principles and have interpretable parameters with a physical meaning
(e.g. Planck's constant, Avogadro's constant). The internals of white-box models are known and can be understood.
Black-box models establish a functional mapping from inputs to outputs by fitting to observations whereby the internals
of the model are irrelevant or unknown. Therefore, the internal parameters of black-box models have no physical meaning
\cite{Sjoberg1995}. Examples of black-box models are non-parametric statistical models (i.e. all kernel-methods
including support vector machines, Gaussian processes, LOESS), neural networks, and tree ensemble methods (e.g. random
forest, gradient boosted trees). Grey-box models also establish a functional mapping from inputs to outputs by fitting
to observations but have only a few parameters and simple interpretable equations.

One possible approach for the identification of grey-box or potentially even white-box models is symbolic regression
(SR) with genetic programming (GP). The main aim of SR is finding well fitting model equations (structure) as well
as their parameters \cite{Koza:1993:pimlssi}. SR offers the possibility of interpretability which can be achieved by
including model complexity as an optimization criterion additionally to model fit. A popular approach for solving SR
problems is GP, which is an evolutionary algorithm that evolves computer programs. It uses a population of solution
candidates (programs) which is iteratively improved by mimicking processes observed in natural evolution, namely
survival of the fittest, recombination, and mutation \cite{Koza1992}. Programs that are better with respect to an
objective function (also called fitness function in GP) are selected with a higher probability for recombination while
bad programs have a low probability to be selected.

SR models can be easily integrated
into mathematical models regardless of the modelling software environment because the commonly used operators and
functions are readily available in most standard libraries. 

SR is particularly suited for the
integration of physical principles and offers the possibility to produce interpretable models which can be integrated
easily into existing software frameworks making it a particularly interesting approach for scientific machine learning~\cite{baker2019workshop} tasks. 

In the present work, we apply SR and GP to extend a constitutive model used to describe the materials response to applied forces, which is a core of numerical metal forming simulations. The main research question thereby is whether such an extended constitutive model achieves accurate enough simulation performance when evaluated with varying process conditions.

\section{Objectives}
The present study is a continuation of the work published in \cite{Kabliman2021} with a goal to improve the constitutive models using GP and SR. The used constitutive model  \cite{Kabliman2019} is white-box and is based on physical laws. It describes the material flow behaviour in terms of internal state variables such as dislocation density. Among the physical values, it contains a number of calibration parameters, which depend on processing conditions (e.g. deformation temperature and strain rate) and are normally fitted to experimental data. To overcome the required regular fitting, the expressions which correlate the calibration parameters to processing parameters should be derived. This task can be addressed using GP and SR. In \cite{Kabliman2021} we have used the  following approach: the calibration parameters were first optimized using a global optimizer, and then the formulas were identified for predicting the optimized values using SR. We call this approach the \emph{explicit} method.

Recently, \citet{Asadzadeh2021} have described a hybrid modelling approach based on symbolic regression whereby the model structure is partially fixed. They have used the approach to extend a physics-based model for a sheet bending process whereby symbolic regression was used to evolve additional terms for the model. They found that partially fixing the structure leads to more consistent symbolic regression results (higher repeatability) and lower model complexity. Additionally, the hybrid modelling approach required only few data points. 

In the present paper, we propose an approach similar to \cite{Asadzadeh2021}, which can evolve the required formulas (extensions) to the constitutive model directly. This may improve the knowledge of the modelled system or process since the functional dependency of calibration parameter values from processing conditions is explicitly established and may be interpreted. We call this approach the \emph{implicit} method. The difference to \cite{Asadzadeh2021} is that we use the approach for a constitutive model which requires simulation to fit stress-strain curves. In the implicit approach the fixed physics-based part is hard-coded in the simulation model and extensions are evolved using multi-tree genetic programming. Additionally, we demonstrate the approach on measured stress-strain curves instead of generated data.

To compare the results from both (explicit and implicit) approaches, we use the same formulation of the used constitutive material model and the same experimental data as in \cite{Kabliman2021}. For better understanding, we also give a detailed description of the (explicit) modelling method used before.

\section{Background}
\subsection{Data Collection}
To generate a data set, a series of hot compression tests 
was conducted for the aluminium alloy AA6082 as described in \cite{Kabliman2019} and \cite{Kabliman2021}. The cylindrical samples (5 mm diameter by 10 mm) were compressed using a deformation and quenching dilatometer DIL805/A from TA Instruments. Compression tests were performed up to a strain of $0.7$ at various temperatures and average strain rates as summarized in Table \ref{tab:training-test}. Invalid measurements at the beginning and end of a test, when the machine resets, were removed. The specimens were induction-heated inside of the dilatometer to the prescribed temperature and afterwards compressed at an average constant strain rate and controlled temperature.  For each set of deformation parameters ($T$ and $\dot{\varphi}$), two identical tests were performed and the average values were calculated. Measurements from hot compression tests are assigned to training and testing sets as shown in Table \ref{tab:training-test}. The training set is itself partitioned into four parts for four-fold cross-validation.

\begin{table}
  \centering
  \caption{\label{tab:training-test}Assignment of hot compression tests to training and testing sets. The training data set is split into four partitions for four-fold cross-validation. The values in each cell indicate the assignment to the partitions.}
  \begin{tabular}{ccccccc}
   &       & \multicolumn{5}{c}{$\dot{\varphi} [1/s]$} \\
   &       & 0.001 & 0.01 & 0.1 & 1 & 10 \\
    \hline
   \multirow{7}{\baselineskip}{\rotatebox[origin=c]{90}{Temperature $[\degree C]$}}
   &   350 & \cellcolor{lightgray} Test   &                            4 &                            3 & \cellcolor{lightgray} Test   &                            2 \\
   &   375 &                            1 & \cellcolor{lightgray} Test   &                            4 &                            3 & \cellcolor{lightgray} Test   \\
   &   400 &                            2 &                            1 & \cellcolor{lightgray} Test   &                            4 &                            3 \\
   &   425 & \cellcolor{lightgray} Test   &                            2 &                            1 & \cellcolor{lightgray} Test   &                            4 \\
   &   450 &                            3 & \cellcolor{lightgray} Test   &                            2 &                            1 & \cellcolor{lightgray} Test   \\
   &   475 &                            4 &                            3 & \cellcolor{lightgray} Test   &                            2 &                            1 \\
   &   500 &                            1 &                            4 &                            3 & \cellcolor{lightgray} Test   &                            2 \\
  \end{tabular}
\end{table}

The stress and strain values were derived during the measurement using the following calculation formulas:

\begin{equation}
 \text{kf}_i = \frac{F_i}{A^{cs}_i},\, 
 A^{cs}_i=\frac{\pi d_0^2 L_0}{4 L_i}, \,
 \varphi_i=\ln{\frac{L_0}{L_i}},\, \dot{\varphi}_i = \frac{d \varphi}{d t}, 
 \label{eqn:data}
\end{equation}
where $i$ indexes subsequent measurements, $\text{kf}_i$ is the stress, $F_i$ is the measured force, $A^{cs}_i$ is the actual cross section of the sample, $\varphi_i$ is the strain, $L_0$ is the initial length at the start of the deformation segment, $L_i$ is the actual sample length, $\dot{\varphi}$ is the strain rate, $dt$ is the time difference between two measurement points, and $d_0$ is the initial sample diameter. 

\begin{figure}
  \centering
  \includegraphics[width=\textwidth]{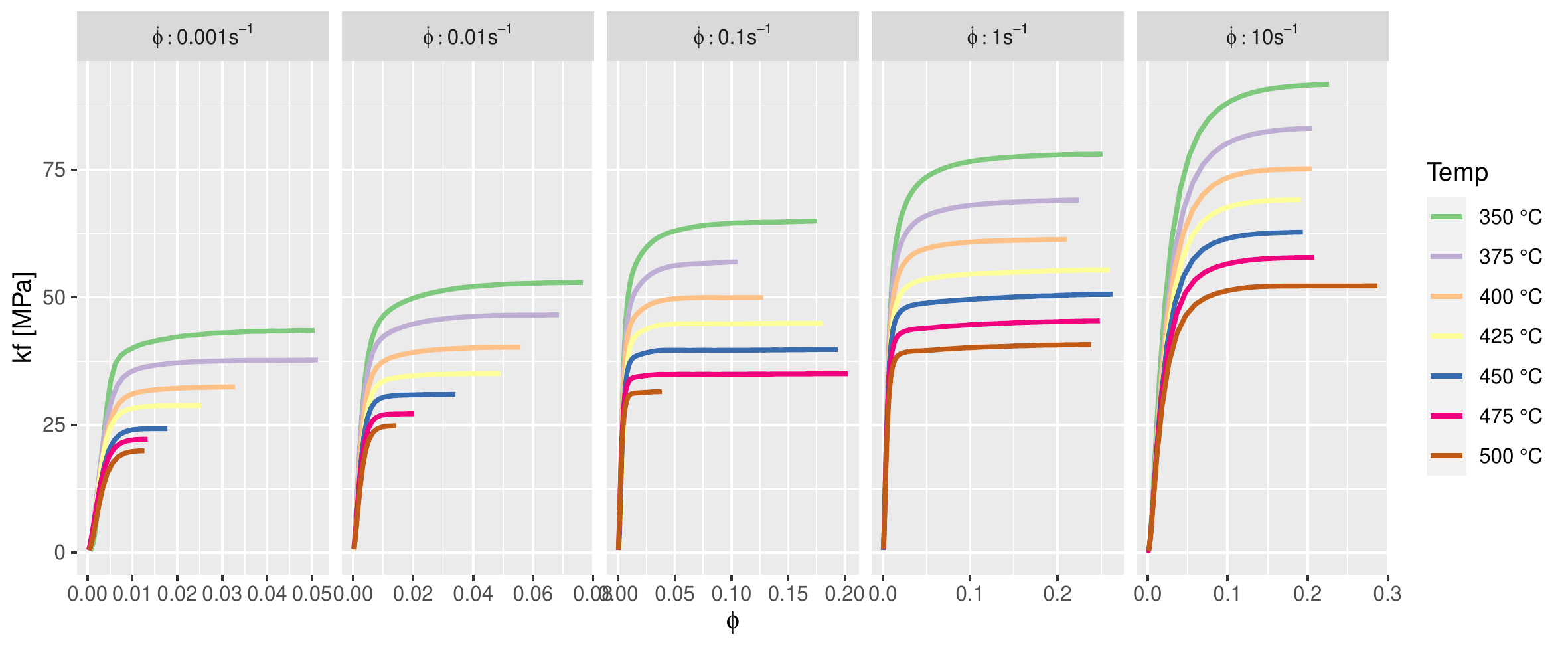}
  \caption{\label{fig:data}Stress-strain curves acquired from the hot compression tests. Only measurements up to the maximum stress value are used for modelling.}
\end{figure}

Figure \ref{fig:data} shows the processed stress-strain data. 

\subsection{Constitutive Model}
\label{sec:model}
Typically, a constitutive model is represented by a sum of an initial yield
(threshold) stress, $\sigma_{y}$, and a strain-dependent part,
$\sigma_{p}$. While the threshold stress is usually a guess value, the
strain-dependent stress might be written as a power law-form or in
terms of material internal state variables. The latter approach is
based on physical laws and, therefore, easily interpretable.
The model used in present work is based on the
evolution of mean dislocation density, $\rho$, and is called Mean Dislocation Density Material Model (MD2M) \cite{Kabliman2019}.

\begin{equation}
  \sigma=\sigma_{y} + MGb\left[\frac{\sqrt{\rho}}{2}+\frac{1}{\delta}\right]
  \label{eqn:model1}
\end{equation}

Here $M$ is the Taylor factor, $G$ is the shear modulus, $b$ is the
norm of the Burgers vector and $\delta$ is a mean sub-grain size. By deformation
at a temperature, $T$, and a strain rate, $\dot{\varphi}$, the
dislocation density will change according to the following equation:

\begin{equation}
  d\rho=\frac{M\sqrt{\rho}}{b A}\dot{\varphi}dt-2BM\frac{d_{ann}}b\rho M\dot{\varphi}dt-2CD\frac{Gb^{3}}{k_{B}T}\left[\rho^{2}-\rho_{eq}^{2}\right]dt
  \label{eqn:model2}
\end{equation}

The first term describes the increase of dislocations and the next two
terms correspond to the dislocations reduction. The first recovery
process happens when two antiparallel dislocations come to a critical
distance, $d_{ann}$. The second recovery process is
thermally-activated and controlled by a self-diffusion along the
dislocations, $D$. When the processing conditions allow, the material
can recover down to an equilibrium state described by the equilibrium
dislocation density, $\rho_{eq}$.

The model contains several physical parameters (e.g. the Boltzmann
constant, $k_{B}$), which can be found in \cite{Kabliman2021}. 
Besides them, there are three calibration parameters $A, B$ and $C$. 
Their values depend on the material, its state and deformation conditions, but these dependencies are not fully
understood. Therefore, these parameters are normally tuned using the experimental stress-strain curves.

From inspection of the white-box model, it can be determined that 
$A, B$, and $C$ must be positive, because, otherwise, the Eqn. \ref{eqn:model2} has no physical meaning. 
Furthermore, $B$ and $C$ depend on the scale of $A$, since the relative contributions of the terms must be of similar size. 
To simplify the optimization and prediction of the parameters, we use an alternative parameterization $u, v$ and $w$ with
\begin{equation}
  A^{-1}  = \exp(u),\, 
  B = \exp(v) A^{-1},\, 
  C = \exp(w) A^{-1},\, 
\end{equation}
and limit the search space to $u \in [-15, 0],\, v \in [-15, 15]\, w \in [-15, 0 ]$.
The transformation stretches the search space non-linearly, whereby the space becomes exponentially larger as $A, B$ or $C$ approach zero. The value for $A$ cannot reach zero. This guarantees a physically feasible solution. The domain for $u,v$ and $w$ is set based on the range of values, which are plausible for the computation material scientists.

\section{Methods}
\subsection{Quantification of the Model Accuracy}
To measure the accuracy of developed model extensions, we use the sum of mean of squared errors
(SMSE) over all tests for the training and testing sets separately. 
The measurement frequency is the same for tests with different strain rates, which implies that
the data set has a variable number of measurements for each test. A
simulation run produces outputs $\hat{\text{kf}}(\varphi)$ with a much higher resolution
for $\varphi$. From these values we keep only the points with matching
measurements $\text{kf}(\varphi)$ and sum up the squared errors. The
metric for model accuracy is the sum over all tests of the mean of squared
errors (SMSE). This puts equal weight on each test even when
the number of measurements differs over the tests.

\begin{align}
  \text{MSE}(\hat{\text{kf}}, \text{kf}_{t}, \varphi_t) & = \frac{1}{n_t} \sum_{i=1}^{n_t}\left(\hat{\text{kf}}(\varphi_{t,i}) - \text{kf}_{t}(\varphi_{t,i})\right)^2 \label{eqn:mse}\\
  \text{SMSE} & = \sum_{t \in \text{tests}}{\text{MSE} (\hat{\text{kf}}, \text{kf}_{t}, \varphi_t)} \label{eqn:smse}
\end{align}
$\hat{\text{kf}}(\varphi)$ are the filtered points from the simulation
and $\varphi_{t}$ and $\text{kf}_{t}$ the measurements from test
$t$. All vectors have $n_t$ elements. We do not normalize the MSE
values for the tests because the target values are all on the same
scale (see Figure \ref{fig:data}) and we aim to reduce absolute not relative errors of
predictions.

\subsection{Explicit Method: Parameter Optimization and Symbolic Regression}
\label{sub:explicit}
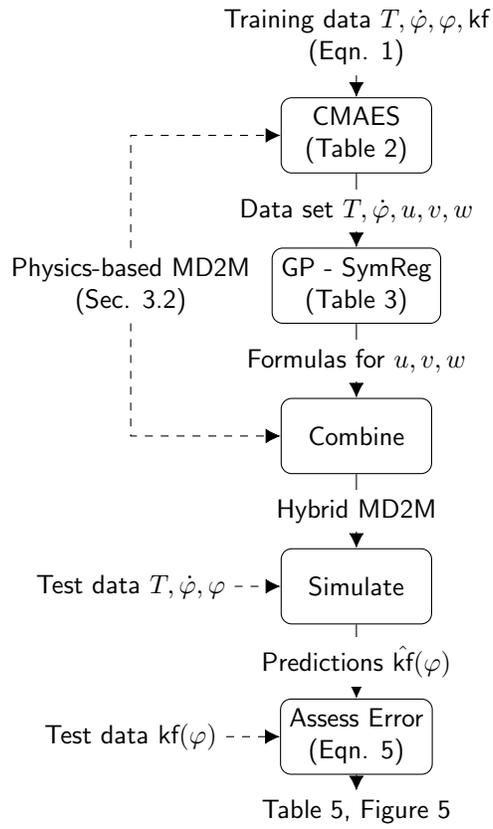
\begin{figure}
  \centering
  \begin{tikzpicture}[node distance=1.5cm,
    every node/.style={fill=white, font=\sffamily}, align=center]
    \node (measure)                                               {Training data $T, \dot{\varphi}, \varphi, \text{kf}$\\(Eqn. \ref{eqn:data})};
    \node (cmaes)    [process, below of=measure, yshift=0.2cm]    {CMAES\\(Table \ref{tab:cmaes})};
    \node (gp)       [process, below of=cmaes, yshift=-0.5cm]     {GP - SymReg\\(Table \ref{tab:symreg})};
    \node (combine)  [process, below of=gp, yshift=-0.5cm]        {Combine};
    \node (model)    [left  of=gp, xshift=-1.5cm]                 {Physics-based MD2M\\(Sec. \ref{sec:model})};
    \node (simulate) [process, below of=combine, yshift=-0.5cm]   {Simulate};
    \node (error)    [process, below of=simulate, yshift=-0.5cm]  {Assess Error\\(Eqn. \ref{eqn:mse})};
    \node (table)    [below of=error,yshift=0.5cm]                {Table \ref{tab:results}, Figure \ref{fig:results-comparison}};
    \node (tstIn)    [left  of=simulate, xshift=-1.5cm]           {Test data $T, \dot{\varphi}, \varphi$};
    \node (tstOut)   [left  of=error, xshift=-1.5cm]              {Test data kf$(\varphi)$};

    \draw[->]             (measure)  -- (cmaes);
    \draw[->]             (cmaes)    -- node[text width=4cm]
                                        {Data set $T, \dot{\varphi}, u, v, w$} (gp);
    \draw[->]             (gp)       -- node[text width=4cm]
                                        {Formulas for $u, v, w$} (combine);
    \draw[->]             (combine)  -- node[text width=4cm]
                                        {Hybrid MD2M} (simulate);
    \draw[dashed,->]      (model)    |- (cmaes);
    \draw[dashed,->]      (model)    |- (combine);
    \draw[->]             (simulate) -- node[text width=4cm]
                                        {Predictions $\hat{\text{kf}}(\varphi)$} (error);
    \draw[dashed,->]      (tstIn)    -- (simulate);
    \draw[dashed,->]      (tstOut)   -- (error);
    \draw[->]             (error)    -- (table);
  \end{tikzpicture}
  \caption{\label{fig:explicit}Workflow of the explicit method for extending the constitutive model (MD2M).
    CMAES is used to optimize parameters $u,v$ and $w$ for each test set from Table \ref{tab:training-test} and
    GP is used to find the expressions for optimized $u,v$ and $w$ depending on $T$ and $\dot{\varphi}$.}
\end{figure}

Figure \ref{fig:explicit} shows the workflow of the explicit
method. A single simulation run produces a stress-strain curve $\hat{\text{kf}}(\varphi)$, which is returned as a table.

First, we use covariance matrix adaptation evolution strategy (CMAES) \cite{hansen2003reducing} to optimize
the calibration parameters for each test in the training set. The
parameters may depend on each other, and we cannot assume that the
optimization problem is convex. This makes parameter fitting difficult \cite{oai:dalea.du.se:2366}. 
Thus, a derivative-free global optimization method such as CMAES or differential evolution can be an
appropriate choice. 
Still, the optimizer may converge to
different solutions. Thus, multiple repetitions
for the same data set are required.
Our goal is to find a function mapping values of the known (processing) parameters to values of the calibration parameters. 
Therefore, it is important that the global optimizer 
converges reliably to the same or similar solutions. Otherwise, we
will fail to find the solution in the subsequent step.
At the end of the optimization, the found solutions are collected into a data set with best values for ($u,
v, w$) for each test.

Next, we use the data set of the optimized ($u,v,w$) values for SR with GP to produce three formulas for the calculation of
$u,v$, and $w$ from the known parameters ($T, \dot{\varphi}$). At this stage, we
recommend to use cross-validation to tune GP parameters (see Table \ref{tab:training-test}). It is
essential to find a GP parameterization that reliably produces a good
solution, because we must select only a single formula for each of
$u,v$, and $w$. After grid-search
for good GP parameters, we execute a GP run with the whole
training set for each of the three calibration parameters and combine
the three formulas with the considered constitutive model (MD2M) to produce the \emph{hybrid} model.
The hybrid MD2M is then used to produce simulation results $\hat{\text{kf}}(\varphi)$ for the
tests in the testing set. 

\subsection{Implicit Method: Evolutionary Extension}
\label{sub:implicit}
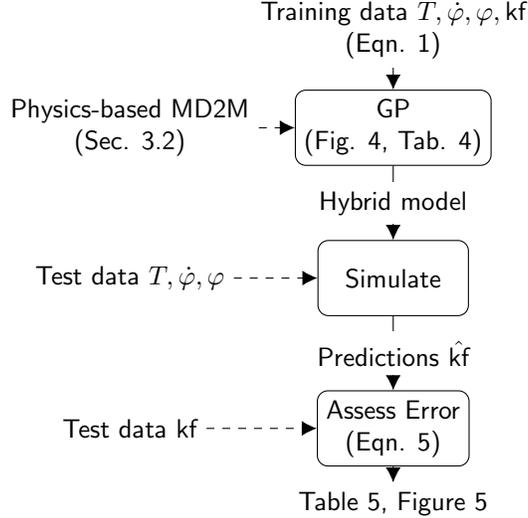
\begin{figure}
  \centering
  \begin{tikzpicture}[node distance=1.5cm,
    every node/.style={fill=white, font=\sffamily}, align=center]
    \node (measure)                                               {Training data $T, \dot{\varphi}, \varphi, \text{kf}$\\(Eqn. \ref{eqn:data})};
    \node (gp)       [process, below of=measure, yshift=0.2cm]    {GP\\(Fig. \ref{fig:gp-alg}, Tab. \ref{tab:gp-parameters})};
    \node (model)    [left  of=gp, xshift=-2.0cm]                 {Physics-based MD2M\\(Sec. \ref{sec:model})};
    \node (simulate) [process, below of=gp, yshift=-0.5cm]        {Simulate};
    \node (error)    [process, below of=simulate, yshift=-0.5cm]  {Assess Error\\(Eqn. \ref{eqn:mse})};
    \node (table)    [below of=error,yshift=0.5cm]                {Table \ref{tab:results}, Figure \ref{fig:results-comparison}};
    \node (tstIn)    [left  of=simulate, xshift=-2cm]           {Test data $T, \dot{\varphi}, \varphi$};
    \node (tstOut)   [left  of=error, xshift=-2cm]              {Test data kf};

    \draw[->]             (measure)  -- (gp);
    \draw[->]             (gp)       -- node[text width=4cm]
                                        {Hybrid model} (simulate);
    \draw[dashed,->]      (model)    -- (gp);
    \draw[->]             (simulate) -- node[text width=4cm]
                                        {Predictions $\hat{\text{kf}}$} (error);
    \draw[dashed,->]      (tstIn)    -- (simulate);
    \draw[dashed,->]      (tstOut)   -- (error);
    \draw[->]             (error)    -- (table);
  \end{tikzpicture}
  \caption{\label{fig:implicit}Workflow of the implicit method for
    extending the constitutive model (MD2M). The extensions 
    are encoded as multi-tree GP individuals and fitness
    evaluation uses simulations of the hybridized model.}
\end{figure}

Figure \ref{fig:implicit} shows the workflow for the implicit method.
In contrast to the
explicit method we do not fit the model to training data by parameter
optimization, but instead use tree-based GP to directly evolve the
three expression trees representing the formulas for the calculation
of $(u,v,w)$. It is important to note that
the output of the MD2M model depends on all three calibration parameters. The
quality of the extended MD2M model can be quantified via the SMSE between simulated and the measured stress-strain curves. However,
we cannot attribute changes in the SMSE (Eqn. \ref{eqn:smse}) to each of the formulas for $(u,v,w)$. Therefore, we use a multi-tree GP for the implicit method.

Each GP individual consists of
three separate trees for the three formulas. For
fitness evaluation the three formulas are used to calculate the
$(u,v,w)$ values for each test. The SMSE
between simulation and measurement over all
tests is used to determine fitness, whereby individuals with smaller
error are assigned a higher fitness value. Therefore, individuals with three
formulas that work well in combination have a higher fitness and are
more likely to be selected.

To highlight how recombination and mutation operations act independently on the
components of the model, we give a pseudo-code for GP algorithm in Figure \ref{fig:gp-alg}.
Crossover between two parent individuals acts on the three components
independently. To produce a child we need to compose three trees from
the parents.  For each of the three components, we first choose
randomly whether a crossover operation should be performed using the
crossover probability parameter. The choice is made independently for
each component. Without crossover we simply select one of the two
trees for this component from the parents randomly. Otherwise a new
tree is created using a sub-tree crossover. This crossover scheme does
not allow exchange of genetic material between separate components. We
recommend a crossover rate smaller than $100$\% to allow combination of
already well working formulas. 
Mutation acts on the three components independently using the mutation
probability parameter.

The individual with highest fitness for the tests in the training set
is selected as the solution. The solution is then used to produce
simulation results $\hat{\text{kf}}(\varphi)$ for the tests in the test
set.

\begin{figure}
  \centering
  \footnotesize
\begin{verbatim}
  Inputs: Model: MD2M(x, theta)  // x: vector of inputs
                                 // theta: parameter vector of length dim
          Matrix of inputs: X = (x_t) // t = 1..tests 
          List of targets:  y = (y_t) // t = 1..tests
                            // each y_t is a time series of measurements
  Output: Hybridized model g(x) = MD2M(x, theta: f(x))

  P = Init(popSize)         // each individual is a vector 
                            // of expressions encoded as trees
  for g = 1 .. maxGenerations
      fitness = [ Evaluate(MD2M, y, X, individual)) 
                  for each individual in P]
      P = order P by descending fitness
      P_next[1] = P[1];     // copy best individual
      for k = 2 .. popSize
          p1 = Select(P)    // select two parents with tournament selection
          p2 = Select(P)
          child = new empty vector of dim expressions 
          for treeIndex = 1 .. dim
              if rand() < crossoverProbability
                  child[treeIndex] = Crossover(p1[treeIndex], p2[treeIndex])
              else
                  child[treeIndex] = rand() < 0.5 ? p1[treeIndex] : p2[treeIndex]

              if rand() < mutationProbability 
                  Mutate(child[treeIndex])
          end               // for all components
          P_next[k] = child
      end                   // for population
      P = P_next
  end                       
  return P[1];              // return individual with best fitness
\end{verbatim}
\caption{\label{fig:gp-alg}Pseudo-code for the multi-tree GP algorithm to
  evolve the extensions for the MD2M model. Individuals contain
  multiple expression trees, one for each element of the parameter
  vector $\theta$. The crossover and mutation act independently on the
  components of individuals.}
\end{figure}

\subsection{Algorithm Configuration}
\subsubsection{Explicit Approach}
For the explicit approach, we execute 30 independent CMAES runs with the parameter settings given in Table \ref{tab:cmaes}. We choose the best values for each of the three parameters ($u,v,w$), which are then back-transformed to produce values for ($A,B,C$). The back-transformed variables are used as target for SR\footnote{We also ran SR experiments to instead predict the transformed parameters $u,v,w$ but found that the results were significantly worse.}. The resulting data set has 35 rows, two input variables, and three target variables. 
We use two subsets of the data for training and testing of models, and assign data from one test completely either to the training set or to the testing set using a systematic partitioning scheme from \cite{Schuetzeneder2020} (see Table \ref{tab:training-test}). 

\begin{table}
  \centering
  \caption{\label{tab:cmaes}CMAES parameters}
  \begin{tabular}{ll}
    Parameter & Value \\
    \hline
    Search space & $u\in [-10.. 0] \times v\in[-15.. 15], \times w\in[-15, 0]$\\
    Generations & 500 \\
    Pop. size & 100 \\
    Initialization & Uniform \\
    Fitness & SMSE (Eqn. \ref{eqn:smse}) for a simulated stress-strain curve\\
    Recombination & Log-weighted \\
  \end{tabular}
\end{table}

The maximum length of symbolic expression
trees is selected using 30 independent repetitions of cross-validation (CV)
with four folds on 24 training samples using the paritions shown in Table \ref{tab:training-test}.
The setting with the smallest median cross-validated root mean squared error (CV-RMSE) is
used to train models on the full training set. The maximum limit for the number of tree nodes is chosen from the set $\{5, 7,
10, 15, 20, 25, 30, 35, 40\}$.  The settings with best CV-RMSE are $25$ nodes for $A$, $35$
nodes for $B$, and $20$ nodes for $C$. These are relatively tight limits for GP, but the grid-search showed that
GP started to overfit with larger models. This can be explained by the small number of data points for training. Models with 20 to 35 nodes (before simplification) are relatively easy to interpret\footnote{In this context it is important to point out that the variable nodes in the leaves of trees implicitly contain a scaling coefficient ($c_i * x_i$). This is counted as only one node in this work while it would be counted as three nodes in most other GP systems.}.

For each of the three targets, we execute 30 GP runs and return
the individual with highest fitness as the solution. The outputs of all GP models are clamped using target-specific limits as shown in Tables~\ref{tab:symreg} and \ref{tab:gp-parameters}. This ensures that the calibration parameter values produced by GP models are physically plausible. For instance, negative parameter values are physically impossible and the maximum values depend on the material. 
We generate 30 hybrid MD2M by combining the three models from the $i$-th SR run for each target and calculate the SMSE for training and testing sets. 

\begin{table}
  \centering
  \caption{\label{tab:symreg}GP parameters for SR as part of the explicit approach.}
  \begin{tabular}{ll}
    Parameter & Value \\
    \hline
    Pop. size & $300$ \\
    Generations & $250$ \\
    Max. length & $A: 25, B: 35, C: 20$\\
    Max. depth & $8$\\
    Initialization & PTC2 \cite{Luke2000}\\
    Selection & Tournament (size $3$)\\
    Recombination & Sub-tree crossover ($90$\% internal nodes)\\
    Mutation & Probability 15\%\\
             & Select randomly: \\
             & For a random parameter: $x \leftarrow x + N(0, 1)$ \\ 
             & For all parameters: $x \leftarrow x + N(0, 1) $\\ 
             & Change the symbol of a random node\\
             & Change a random variable node \\
    Clamp predictions & $A: [0..150], B: [0..150], C:[0..1]$\\
    Fitness & Sum of squared errors (for $A,B,C$ predictions)\\
    Replacement & Generational with one elite. \\
    Function set & $\{+, -, *, \%, \exp(x), \log(x)\}$ \\ 
    Terminals & $50$\% variables, $50$\% numeric parameters \\
              & Variables: \{temp, $\dot{\varphi}, \log_{10}(\dot{\varphi}$)\} \\
              & Numeric parameters $\sim U(-20, 20)$ \\
  \end{tabular}
\end{table}

\subsubsection{Implicit Approach}
We use tree-based GP with generational replacement
with elitism. As described
above, each individual contains three trees ($u, v, w$). Trees are limited to 25 nodes and
a maximum depth of 10 for each component.  Our GP system
initializes trees randomly using PTC2, whereby for each leaf it first
randomly determines the leaf type: variable or parameter. All
variable nodes always include a scaling factor sampled randomly from
$N(0, 1)$\footnote{This is the default of the GP system used. This increases the number of parameters in the SR models and can be helpful or detrimental for fitting. We have not analysed the effects of removing scaling parameters for variables.}. The parameters are sampled randomly from $U(-20, 20)$. The GP parameter settings used for the implicit approach are given in Table \ref{tab:gp-parameters}.

\begin{table}
  \centering
  \caption{\label{tab:gp-parameters}GP parameter settings for the implicit approach.}
  \begin{tabular}{ll}
    Parameter & Value \\
    \hline
    Pop. size & $5000$ \\
    Generations & $250$ \\
    Max. length & $25$\\
    Max. depth & $10$\\
    Initialization & PTC2 \\
    Selection & Tournament (size $7$)\\
    Recombination & Probability $30$\% for each tree of an individual\\
                  & Sub-tree crossover ($90$\% internal nodes)\\                  
    Mutation & Probability $15$\% for each tree of an individual\\
             & Select randomly: \\
             & For a random parameter: $x \leftarrow x + N(0, 1)$ \\ 
             & For all parameters: $x \leftarrow x + N(0, 1) $\\ 
             & Change the symbol of a random node \\
             & Change a random variable node \\
    Clamp predictions & $u: [-15..0], v: [-15..15], w:[-15..0]$\\
    Fitness & SMSE (Eqn. \ref{eqn:smse}) \\
    Replacement & Generational with one elite. \\
    Function set & $\{+, -, *, \%, \exp(x), \log(x)\}$ \\
    Terminals & 50\% variables, 50\% numeric parameters \\
              & Variables: \{temp, $\dot{\varphi}, \log_{10}(\dot{\varphi}$)\} \\
              & Numeric parameters $\sim U(-20, 20)$ \\
  \end{tabular}
\end{table}

\section{Results}
\label{sec:results}

\begin{figure*}
  \centering
  \includegraphics[height=3.5cm]{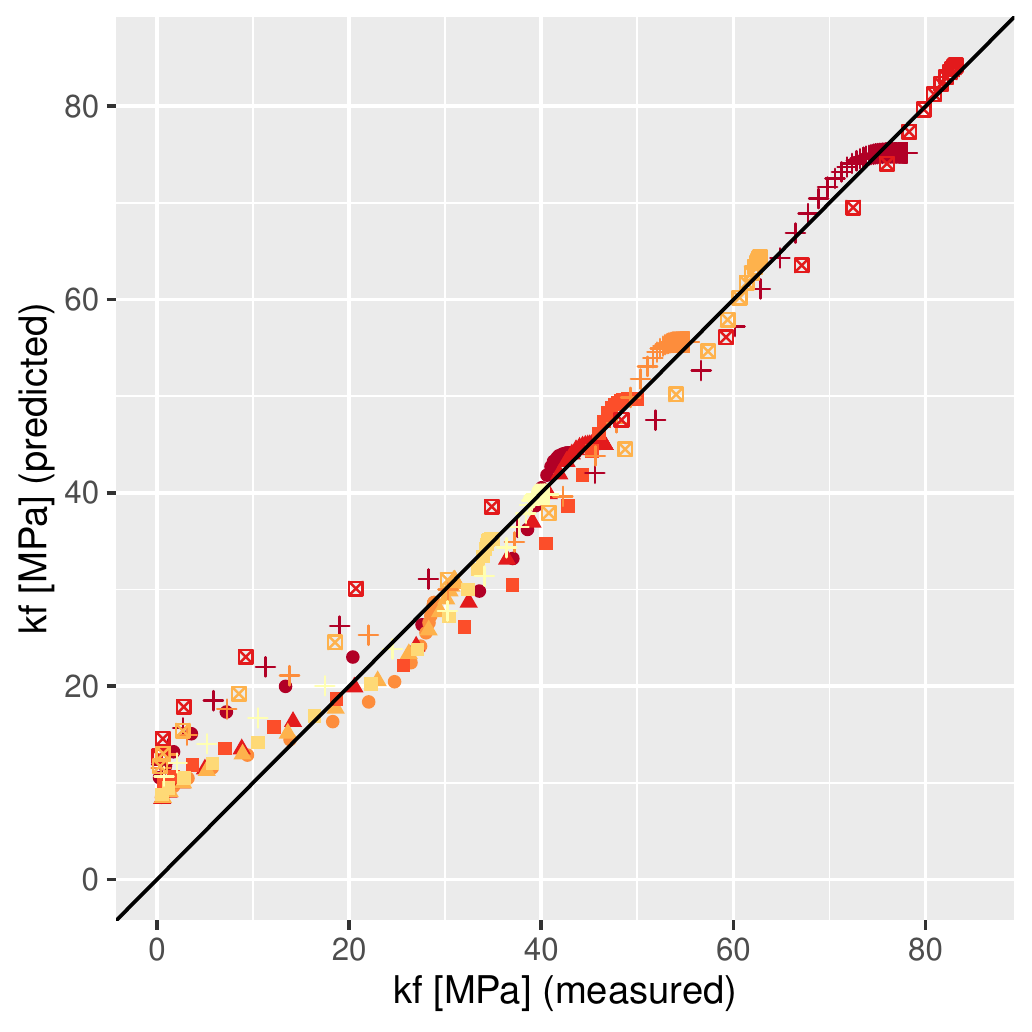}
  \includegraphics[height=3.5cm]{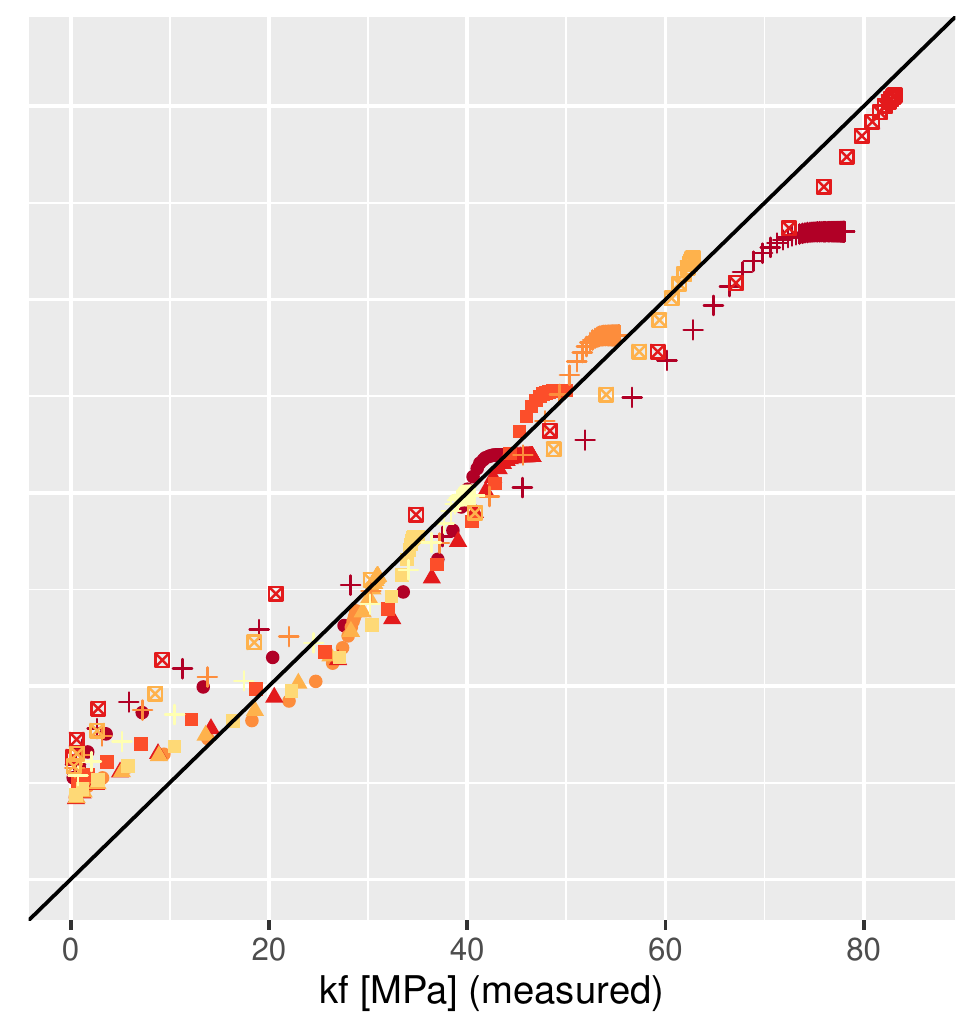}
  \includegraphics[height=3.5cm]{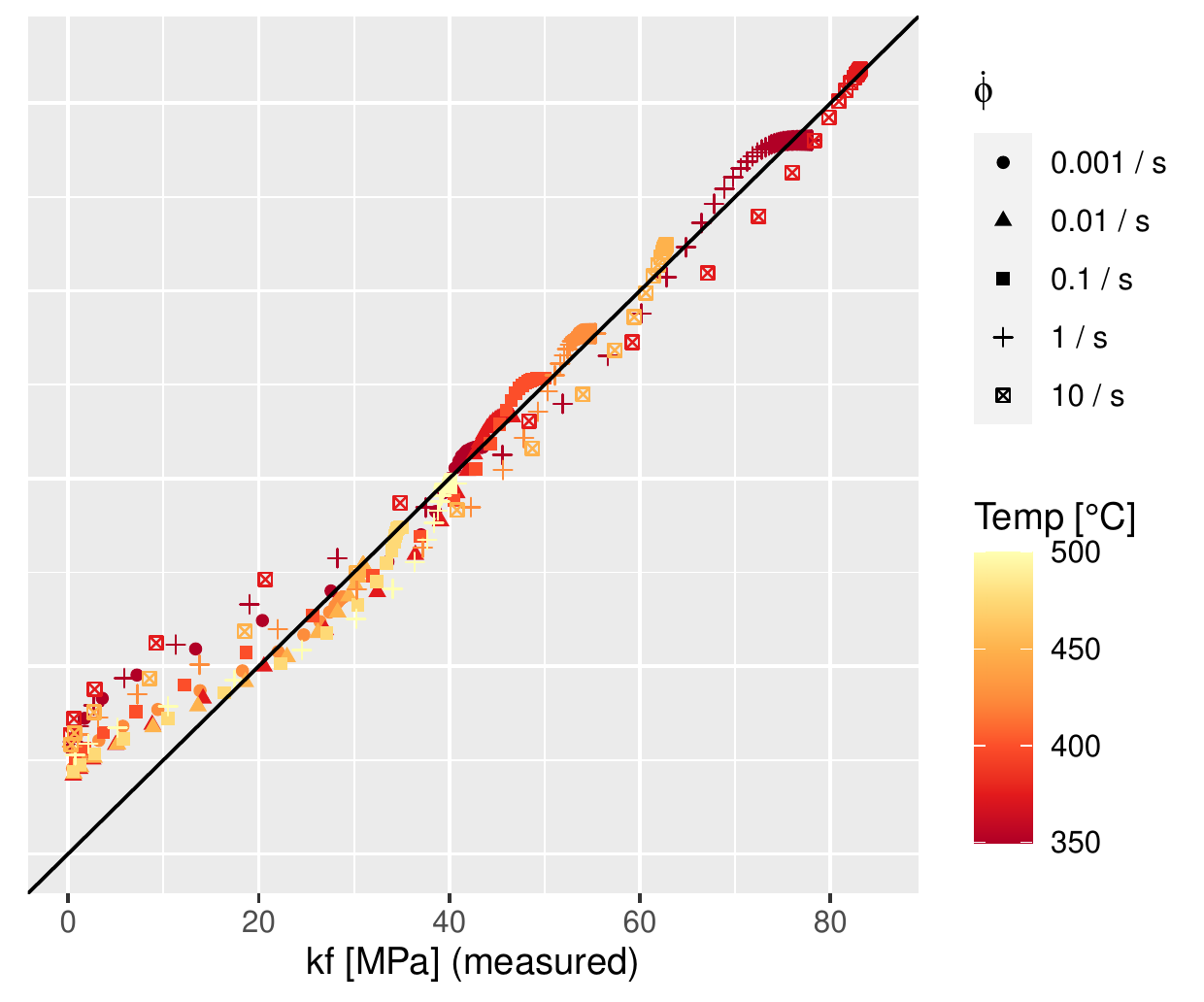}
  \caption{\label{fig:results-comparison}Scatter plots for predictions over measurements for interpolation (left),  the explicit (middle), and the implicit (right) approach. The implicit approach produces a better fit for high stress values.}
\end{figure*}

Figure \ref{fig:results-comparison} visualizes the simulation outputs when A,B,C are calculated via interpolation, the implicit, and the explicit model.
Only the 11 hot compression tests from the testing set are shown.
For the implicit and the explicit model, the model with best training SMSE from the 30 runs is used.
A systematic deviation for small kf values is apparent for all methods, which is due to a systematic bias of the MD2M. It is not possible to calibrate the MD2M to improve the simulation results for small kf values even with CMAES. 

\begin{figure}
  \centering
  \includegraphics[width=\columnwidth]{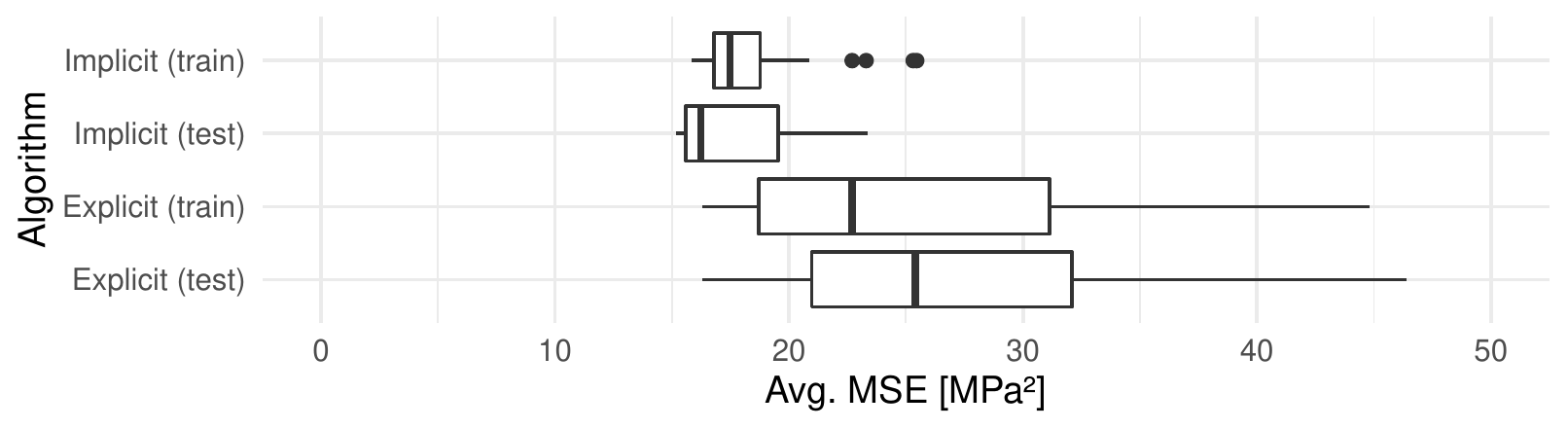}
  \caption{\label{fig:boxplot}Box-plots for the average MSE over 30 repetitions for training and testing sets. 
    Not shown: twelve outliers for the explicit method (training: $2$, test: $10$).}
\end{figure}

Figure \ref{fig:boxplot} illustrates box-plots for the average training and test MSE values achieved by both methods over 30 repetitions. The implicit method has smaller MSE values on average on the training and testing sets. Additionally, the variance for the implicit method is much smaller than for the explicit method. The non-parametric Brown-Mood median test (for samples with different variances) indicates to reject the null-hypothesis of equal medians for the training set (p-value = $0.0001792$) and for the testing set (p-value = $1.183\cdot 10^{-07}$). For the optimized parameter values (CMAES) the average test MSE is $13.58\, \text{MPa}^2$ (RMSE=$3.69\, \text{MPa}$), for the interpolated values it is $14.35\, \text{MPa}^2$ (RMSE=$3.79\, \text{MPa}$).

The median MSE values for each test for the hybrid models produced by 30 independent runs of the explicit and implicit method are collected in Table \ref{tab:results}. The implicit method has smaller MSE values than the explicit method for the training and the testing sets. For comparison, the median MSE values achieved with optimized parameters (CMAES) and with interpolation are given as a reference. The CMAES runs reliably converged to the same solutions and give an indication of the best possible MSE values that can be achieved when using the MD2M.

Instead of learning a model we can also interpolate A,B,C from the optimized parameters for the compression tests in the training set. We found the best results with linear interpolation along the temperature dimension. For this we calculate the parameters for each test from the values found for lower and higher temperatures. The results of linear interpolation are shown in the interpolation column in Table \ref{tab:results}.

Summarizing the results, we found a better model (with on average a smaller SMSE) using the implicit method, than with the explicit method. With increasing strain rate, the fit achieved with the explicit method gets better. For the highest strain rate, the explicit method produces a better fit for 5 of 7 tests. Linear interpolation works well and produces even better results than the implicit method. However, interpolation does not give us a simple formula to calculate A,B,C.

\begin{table}
  \centering
  \footnotesize
  \caption{\label{tab:results}Median MSE values over 30 runs for the
    explicit and implicit method for the training and
    testing sets. The MSE values for CMAES and interpolation are shown as a reference value as they are indicative of the best achievable value 
    for the MD2M.}
  \begin{tabular}{@{}
      c
      >{$}r<{$}
      >{$}r<{$}
      >{$}r<{$}
      >{$}r<{$}|
      >{$}r<{$}
      >{$}r<{$}
      @{}}    
     & \text{Temp} & \dot{\varphi} & \text{CMAES} & \text{Interp.} & \text{Explicit} & \text{Implicit} \\    
     & [\degree C] &  [s^{-1}] & \text{MSE}\, [\text{MPa}^2] & \text{MSE}\, [\text{MPa}^2] & \text{MSE}\, [\text{MPa}^2] & \text{MSE}\, [\text{MPa}^2] \\    
    \hline
    \multirow{24}{\baselineskip}{\rotatebox[origin=c]{90}{Training}}
                               & 375 &  0.001  &  9.44 &  9.71 & 22.23 & \mathbf{11.48} \\
                               & 400 &  0.001  & 12.70 & 12.95 & 21.11 & \mathbf{14.63} \\
                               & 450 &  0.001  & 14.99 & 15.81 & 28.55 & \mathbf{18.31} \\
                               & 475 &  0.001  & 14.04 & 14.46 & 33.87 & \mathbf{17.19} \\
                               & 500 &  0.001  & 14.94 & 16.69 & 26.34 & \mathbf{21.54} \\
                               & 350 &  0.01   &  7.76 & 11.79 & \mathbf{10.44} & 10.58 \\
                               & 400 &  0.01   &  6.23 &  6.20 & 10.70 &  \mathbf{8.33} \\
                               & 425 &  0.01   &  8.38 &  8.66 & 12.02 &  \mathbf{9.64} \\
                               & 475 &  0.01   & 10.71 & 10.96 & 13.92 & \mathbf{12.22} \\
                               & 500 &  0.01   & 13.37 & 13.87 & 23.63 & \mathbf{16.20} \\
                               & 350 &  0.1    &  7.88 &  8.13 & 10.41 &  \mathbf{9.79} \\
                               & 375 &  0.1    &  9.61 & 17.50 & 14.73 & \mathbf{11.56} \\
                               & 425 &  0.1    &  6.31 &  8.02 &  8.65 &  \mathbf{8.03} \\
                               & 450 &  0.1    &  3.85 &  3.95 &  5.66 &  \mathbf{4.79} \\
                               & 500 &  0.1    &  7.88 &  8.74 &  \mathbf{9.63} & 11.09 \\
                               & 375 &   1     &  7.82 &  8.19 & 11.27 & \mathbf{10.04} \\
                               & 400 &   1     &  6.54 &  7.01 &  \mathbf{7.88} &  7.94 \\
                               & 450 &   1     &  4.89 &  4.95 &  \mathbf{5.25} &  7.29 \\
                               & 475 &   1     &  4.45 &  4.57 &  \mathbf{4.76} &  6.57 \\
                               & 350 &  10     & 43.58 & 45.70 & 50.92 & \mathbf{50.81} \\
                               & 400 &  10     & 38.07 & 38.45 & \mathbf{38.54} & 38.57 \\
                               & 425 &  10     & 38.86 & 39.11 & \mathbf{39.14} & 40.08 \\
                               & 475 &  10     & 31.05 & 31.24 & \mathbf{31.41} & 33.53 \\
                               & 500 &  10     & 18.59 & 19.15 & \mathbf{20.49} & 22.60 \\
                        & \text{Avg} &         & 14.25 & 15.24 & 19.23 & \mathbf{16.78} \\
    \hline    
    \multirow{11}{\baselineskip}{\rotatebox[origin=c]{90}{Test}}
                               & 350 & 0.001 & 17.07 & 18.39 & 41.24 & \mathbf{21.59} \\
                               & 425 & 0.001 & 14.43 & 15.24 & 21.81 & \mathbf{16.76} \\
                               & 375 &  0.01 &  5.92 &  6.73 &  9.38 &  \mathbf{9.21} \\
                               & 450 &  0.01 &  8.99 &  9.15 & 11.44 & \mathbf{10.50} \\
                               & 400 &   0.1 &  5.07 &  5.94 &  \mathbf{5.83} &  6.43 \\
                               & 450 &   0.1 &  2.38 &  2.51 &  \mathbf{2.86} &  3.53 \\
                               & 350 &     1 &  9.94 & 12.42 & 81.95 & \mathbf{15.11} \\
                               & 425 &     1 &  6.12 &  7.00 &  8.52 &  \mathbf{7.66} \\
                               & 500 &     1 &  3.78 &  4.00 &  \mathbf{4.68} &  6.22 \\
                               & 375 &    10 & 42.98 & 43.50 & 45.62 & \mathbf{44.47} \\
                               & 450 &    10 & 32.77 & 32.93 & \mathbf{32.95} & 34.65 \\
                        & \text{Avg} &       & 13.59 & 14.35 & 24.21 & \mathbf{16.01} \\
  \end{tabular}
\end{table}

Equations \ref{eqn:u}, \ref{eqn:v}, \ref{eqn:w} show the resulting set of
expressions for $u, v$ and $w$ produced using the implicit
method and having the best performance on the test set after algebraic
simplification. To determine $A, B$ and $C$, the back-transformation as shown in
Equations \ref{eqn:a}, \ref{eqn:b}, \ref{eqn:c} is required, whereby $\text{clamp}(x, l, u)$ returns $\min(u, \max(l ,
x))$. The models identified by GP are non-linear in the input
variables ($T\,, \dot{\varphi}$) and very short. Most of the parameters
are linear which facilitates interpretation of the formula.  The
formulas identified by the explicit approach found in \cite{Kabliman2019} had a similar complexity,
because the tree size restrictions for both algorithms were similar.

\begin{align}  
  u(T,\dot{\varphi}) = & -\log(0.11 T + 3.734 \dot{\varphi} - 17.34) - 0.069 \log(\dot{\varphi}) \label{eqn:u}\\
  v(T) = & -4.35 \log(T) + 4.938 \label{eqn:v}\\
\begin{split}
  w(T, \dot{\varphi}) = & -13.708 \log(\dot{\varphi})T^{-1} - 10205 \log(\dot{\varphi}) T^{-2}+13675 T^{-2} \\
  & + 0.777 \log(\dot{\varphi}) - 0.8657
\end{split}\label{eqn:w} \\
  A(T,\dot{\varphi}) = &\exp\left(-\text{clamp}(u(T,\dot{\varphi}), -15, 0)\right) \label{eqn:a} \\
  B(T,\dot{\varphi}) = &\exp\left(\text{clamp}(v(T), -15, 15)\right) A(T,\dot{\varphi})^{-1} \label{eqn:b} \\
  C(T,\dot{\varphi}) = &\exp\left(\text{clamp}(w(T,\dot{\varphi}), -15, 0)\right) A(T,\dot{\varphi})^{-1} \label{eqn:c}
\end{align}

\begin{figure}\centering
  \includegraphics[width=0.7\textwidth]{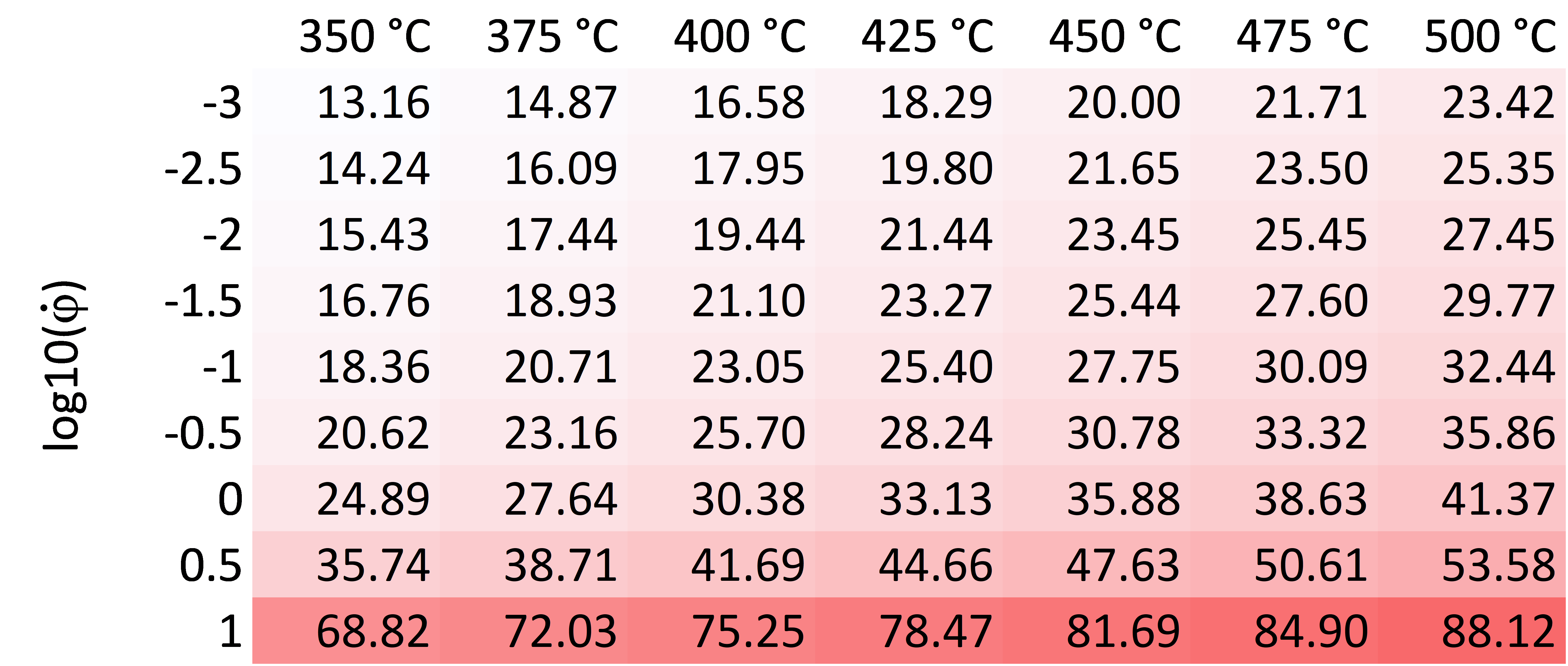}\\[10pt]
  \includegraphics[width=0.7\textwidth]{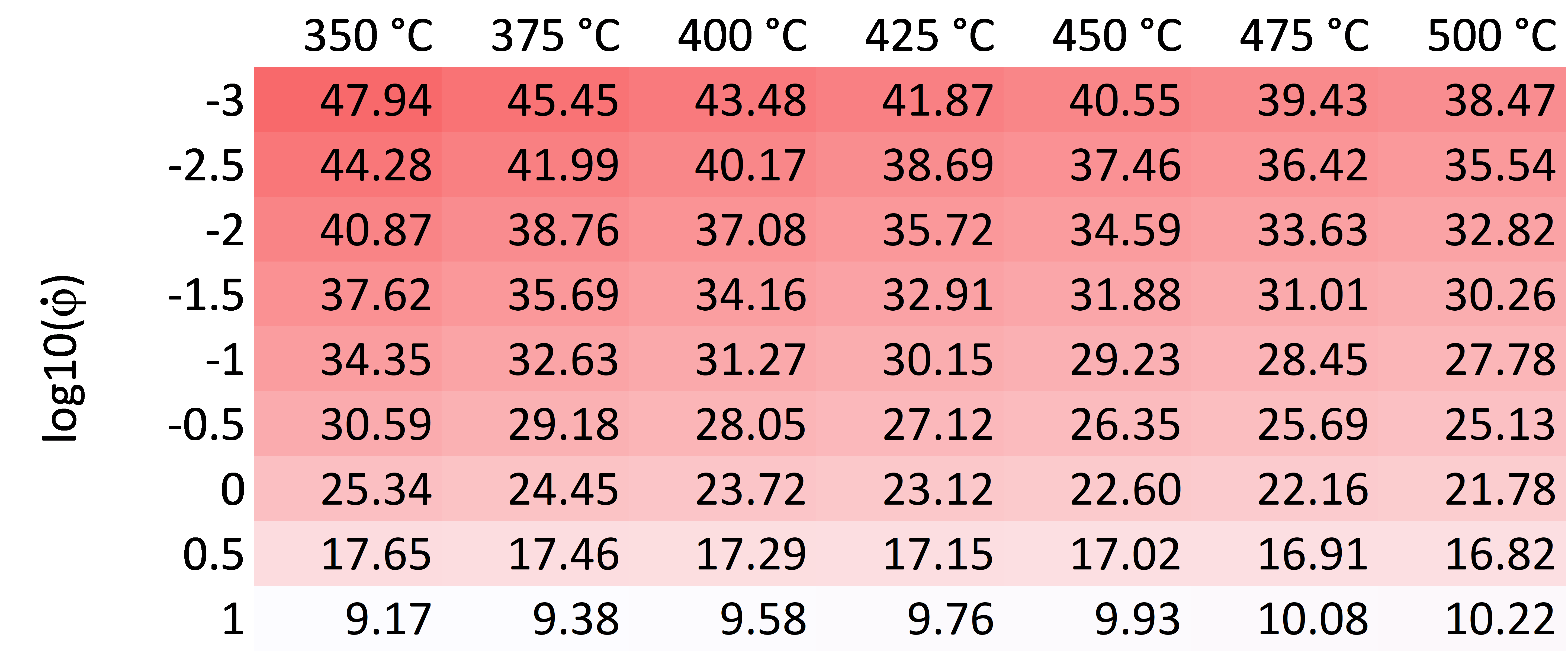}\\[10pt]
  \includegraphics[width=0.8\textwidth]{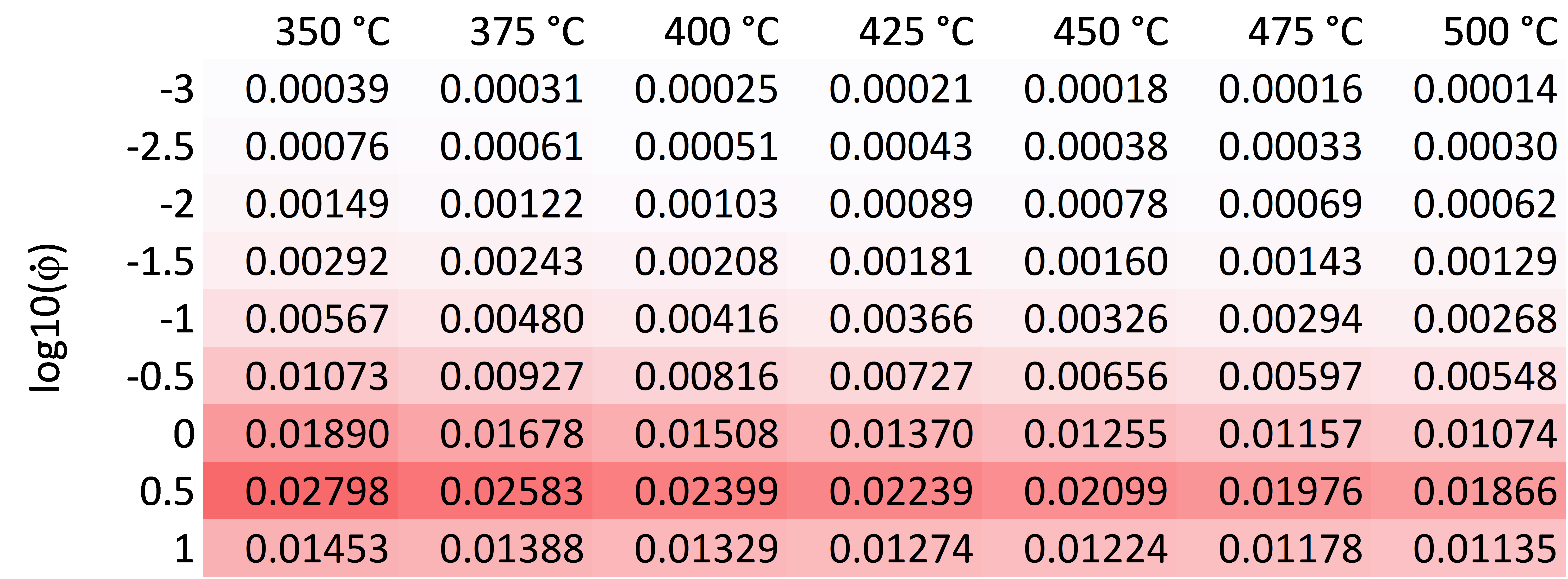}
  \caption{\label{fig:tables}Predicted values for the calibration parameters A, B, and C.}
\end{figure}

Finally, the predicted ($A, B, C$) values for the tested range of temperatures and strain rates are shown in Figure \ref{fig:tables} to visualize the correlation with the processing parameters.

\section{Discussion}
The goal of the present study was to extend the physics-based constitutive model using GP and SR for predicting of unknown relations of the calibration parameters to the processing conditions. We showed that it is possible to evolve short formulas using two (explicit and implicit) methods. However, there are several caveats that we want to discuss below and plan to address in future work.

\subsection{Comparison to Classical Interpolation Methods}
Instead of finding a model for the calculation of the calibration parameters it is possible to use linear or cubic spline interpolation. In fact, linear interpolation along the temperature dimension produced better fitting results than the model. However, the equations produced by GP are more attractive compared to interpolation, since they have only a few parameters and are easily interpretable. For example the interpolation map for parameter $A$ has 24 coefficients while the best model (Eqn: \ref{eqn:u}) has only four coefficients. The expressions for the three calibration parameters can be combined easily with the expressions of the M2DM. 

\subsection{Runtime and Convergence}
For the explicit method, the runtime for one CMAES run was
approximately $6$ hours (10 minutes for each of the 35 tests) and the
runtime for SR was just a few minutes, as there were only 24 training
points. Linear interpolation is computationally very cheap therefore much faster than SR, but it also requires the same optimization runs with CMAES.
The implicit method is computationally much more expensive because each GP fitness evaluation requires a separate simulation run. The runs took approximately 45 days on average single-threaded.
Concerning runtime the explicit method is therefore much better than the implicit approach. The implicit method however provides more options for the evaluation of the fit and led to significantly more accurate models. The runtime can be reduced through parallelization of the simulation runs.
We cannot definitively state whether more runtime for the explicit approach
would improve the results, because longer runs were not tested. 
However, we analysed the convergence of CMAES runs and observed
that it reliably converged to the same points with the parameters
that we have chosen. Therefore, we believe it is unlikely that much
larger population sizes or more generations would have improved the
results for CMAES. The runtime of CMAES is the main
contributor to the total runtime for the explicit approach so
increasing the number of generations for SR would increase total
runtime only by a small percentage.

\subsection{Combination of Models}
The implicit approach directly evolves a set of models that produces a
good prediction in combination, while the explicit approach evolves each
model independently. To create 30 hybrid models for evaluation, we
used the three models from the first run for the first combined model,
the three models from the second run for the second combined model and
so on. The finally selected combined model is the combination with the
best performance (SNMSE) on the training set. However, from the 3*30
individual models, we could also create all 90 combinations with minor
overhead. This could potentially increase the chance to build and
select a good combination of models, but might also increase the chance
for overfitting.

\subsection{Choice of Fitness Function for SR}
One potential reason for the worse performance of the explicit method is that
the chosen fitness function for SR could be inappropriate. The implicit
method uses the fit for the simulated stress-strain curve as a
fitness criterion and therefore directly optimizes the error that we measure
for the test set. With the explicit method, we build formulas to minimize the
error to the $(u,v,w)$ values produced by CMAES. In our experiments, we used the
sum of squared errors as the fitness function for this SR step based on the assumption that
a formula with smaller squared error for $(u,v,w)$
also leads to a better fit for the simulated stress-strain
curve $\hat{\text{kf}}(\varphi)$. Based on our results, we believe that this assumption does not
hold.  For example, it may be necessary to predict a value more
accurately as it approaches zero, while for larger target values we may
also allow larger errors. In this case, it could be better to minimize
the relative error. This can be considered an advantage of the
implicit approach, because it frees us from the burden to select an
appropriate fitness function for the intermediate SR step. It would be
worthwhile to study different fitness functions in future work to gain
a better understanding about the effect of this choice on the overall
performance of the approach.

\subsection{Implementation Effort}
The explicit method is easier to implement than the implicit method,
because the separate steps in the workflow only require to call
readily available and well-tuned software components (CMAES, SR
tools). For the implicit method, it is necessary to use a
multi-tree GP system or to adapt a GP system accordingly. Additionally,
for the fitness evaluation, the simulation model has to be implemented or
connected to the GP system. Most GP systems, however, allow this kind of
extension.

\section{Related studies}

The well-maintained GP bibliography \cite{langdon:2019:GPEM} contains
many references to prior work, in which GP has been used for
constitutive modelling in particular for predicting stress for
various materials. GP for multi-scale material
modelling is extensively discussed in \cite{Sastry:thesis}.
Two early works which describe the application of GP for the
identification of constitutive models are \cite{Schoenauer1996} and \cite{SebSch97}.
The approach described in \cite{Schoenauer1996} is
especially notable as it uses specific operators to ensure that the
resulting models have physical interpretation as elastic, plastic and
viscous components.

Since then GP has been used extensively for constitutive modelling such as
for modelling flow stress for various metallic materials \cite{Brezocnik:2000:JTP, Brezocnik:2001:MPT,Brezocnik:2002:JIM} including aluminium alloys \cite{Sastry:2004:ijmsce},
for modelling stress
distribution in cold-formed copper alloys \cite{Brezocnik:2004:IJAMT} and X6Cr13 steel \cite{brezocnik:2005:MMP},
for predicting impact toughness
\cite{Gusel:2006:CMS},
for the identification of visco-elastic models for rocks \cite{Feng:2006:IJRMMS},
and stress-strain behaviour of sands under cyclic loading \cite{Shahnazari:2010:EG, Shahnazari:2014:JCEM},
for predicting material properties under hot deformation, in particular for carbon silicon steel \cite{kovacic:2005:MMP},
and a nickel-based alloy \cite{Lin:2017:Vacuum},
for predicting the presence of cracks in hot rolled steel \cite{Kovacic:2011:QUALITY},
for modelling tensile strength, electrical conductivity of cold-drawn copper alloys \cite{Gusel201115014}, 
for prediction of shear strength of reinforced concrete beams \cite{Gandomi:2014:MS},
for formulating the stress-strain relationship of materials in \cite{Gandomi:2015:IPSE},
and for predicting fatigue for 2024 T3 aluminium alloys under load \cite{Mohanty:2015:ASC}.
GP has further been used for predicting non-linear
stress-strain curves (e.g. for aluminium and stainless steel alloys)
\cite{Kabliman2019,Cevik:2007:ES,Schuetzeneder2020}, predicting elastic distortional buckling stress
of cold-formed steel C-sections \cite{Pala20081495}, predicting residual stress in
plasma-nitrided tool steel \cite{Podgornik:2011:MMP}, and modelling
mechanical strength of austenitic stainless steel alloys (SS304) as a
function of temperature, strain and strain rate
\cite{Vijayaraghavan:2017:Measurement}.

An evolutionary method for polynomial regression and the combination
with finite element analysis has been used for constitutive modelling
and applied for instance to predict the behaviour of soils under
drained and undrained load conditions in
\cite{FULL_THESIS_MRezania2008} and
\cite{Ahangar-Asr:thesis}. Furthermore, evolutionary algorithms have been used for the optimization of
calibration parameters of constitutive models in
\cite{DOI:10.1504/IJMMS.2011.043078} and optimizing alloy composition
using Gaussian process surrogate models and constitutive models for
simulation \cite{oai:hal.archives-ouvertes.fr:hal-00961230}.  In
\cite{Mulyadi2006}, different methods for parameter optimization of a
constitutive model for hot deformation of a titanium alloy have been
tested and compared with the predictions made by artificial neural
networks.

All of the papers discussed above describe a from of regression modelling. In those papers, GP is used for
supervised learning to establish a free-form constitutive model using
SR (e.g. \cite{Sastry:thesis},
\cite{Schoenauer1996}) or alternatively the model structure is fixed and parameters are optimized using
evolutionary algorithms.
We are, however, mainly interested in combining or extending physics-based models with
machine learning models and found only a few papers with a similar focus in the material science domain.

A hybrid modelling approach using a physics-based
model and neuro-fuzzy evolution has been described and applied for modelling
thermo-mechanical processing of aluminium alloys in
\cite{Abbod2002}. The same authors later sketch a similar GP-based approach
in \cite{abbod2007}. This work is similar to the present study, but there are
several important differences. \citet{abbod2007} used a simpler physics-based model
for flow stress and predicted only three relevant points in the stress-strain curves (steady-state flow stress, the relaxation stress, and the relaxation strain) instead of the full curve. The authors first fit neuro-fuzzy models to the data and only later used GP to find short equations that predict the output of those models. In contrast to our approach proposed in the present work, the resulting GP models are not directly linked to the physics-based constitutive model. Instead, the pre-calculated features and sub-expressions were derived from the physics-based model within GP to produce similar expressions. 

\citet{Versino2017} have described different methods for physics-informed
SR for modelling of the stress-strain curves. The
methods include addition of artificial data points to improve
extrapolation, constraints (e.g. to force models to be non-negative), seeding of a GP population with initial solutions based on the physics-based models, and user-defined features using building blocks derived from the physics-based models (e.g. non-linear transformations of input variables). Seeding GP with the physics-based model is similar to our approach but does not guarantee that the evolved model has the same structure as the physics-based model. In the conclusions \citet{Versino2017} state: \emph{[...] model development can [again] be expertly guided by choosing appropriate building blocks, avoiding functions that might introduce excessive numerical issues. At the same time [...] symbolic regression presents clear limits. When no experimental data or expert knowledge is available, the behaviour of obtained models is highly unpredictable, and unlikely to be rooted in solid physics. [...] Moreover, symbolic regression will probably return completely different models for different materials, limiting the re-usability of a result. Additionally, as EAs are stochastic in nature, there is no guarantee that two runs of the algorithm on the same dataset will provide exactly the same results, introducing reproducibility problems.} We try to partially alleviate these issues by extending the physics-based model with GP. This ensures that at least the core of the model remains unchanged.

\citet{Sedighiani2020} have described a computationally efficient method for the identification of constitutive parameters from stress-strain curves and demonstrated the method for a number of models including a dislocation-density-based crystal plasticity model for steel. They use a genetic algorithm to fit the parameters and response surface models as surrogate models to save simulations and reduce runtime. In contrast to the method proposed in this paper, the method does not produce formulas which relate the constitutive parameters to the load parameters. Instead it produces the parameter values directly. The method could be used to speed-up the first step (parameter identification) within the explicit approach described below.

\section{Conclusions}
We used machine learning for extending a physics-based constitutive model, which describes the materials behaviour in terms of internal state variables. Using GP and SR, we could derive three short expressions for the unknown dependencies of the model calibration parameters to known impact variables such as the processing conditions. As a result, hybrid physics-based and data-driven constitutive models were formulated. We compared two, explicit and implicit, methods for derivation of required formulas. 
 
The critical step in the explicit method is the combination of the SR models for the calibration parameters with the physics-based model. Even though the individual SR models were able to predict the calibration parameters accurately 
for training and testing partitions, they did not perform well in combination. The implicit method instead directly optimizes the fit of the hybrid material model, and is able to evolve a combination of short formulas. Thus, the results have shown that the hybrid material models produced by the implicit method have significantly better predictive accuracy on average than the models produced by the explicit method. Additionally, the implicit method has a smaller variance. However, the results achieved with traditional linear interpolation are better than the results obtained by the two model-based approaches.

We recommend the implicit approach over the explicit approach for finding formulas. It allows the end-to-end
fitting of the simulation model and does not have multiple
intermediate steps of model selection, where an appropriate fitness
function has to be chosen. However, the implicit approach comes at a
significantly greater computational expense, because more evaluations
of the simulation model are required. In particular, the computational demand is much higher compared to traditional interpolation methods. To improve the runtime, future
work could try to either improve the explicit approach or try to
incorporate surrogate models instead of the simulation model. However, even with those improvements it will not be possible to reach similar runtimes as traditional interpolation methods. 

The use of the formulas instead of the calibration parameters might help to overcome the required regular fitting and reduce the amount of necessary measurements.
Moreover, the main advantage of the proposed approach is that parameters A, B, and C directly depend on the processing conditions which gives a higher resolution of the calibration parameter values. The accuracy of the calculated local flow stress is therefore higher compared to fixed calibration values.
The proposed approach is of a general purpose and can be applied in other areas, when the relation of model parameters to impact factors is not well understood. 

\subsection*{Acknowledgements}
The authors gratefully acknowledge support by the Christian Doppler Research Association and the Federal Ministry of Digital and Economic Affairs within the Josef Ressel Center for Symbolic Regression. This work was supported by the Federal Ministry for Climate Action, Environment, Energy, Mobility, Innovation and Technology (in German, BMK) in a framework of the project LiMFo (Light Metal Forming).